\documentclass[11pt,a4paper]{article}
\usepackage[hyperref]{acl2021}
\usepackage{times}
\usepackage{latexsym}

\usepackage{microtype}

\usepackage{amsmath}
\usepackage{amsfonts}
\usepackage{amssymb}

\usepackage{booktabs}
\usepackage{multirow}
\usepackage{multicol,lipsum,microtype}

\usepackage{diagbox}

\usepackage{epsfig}
\usepackage{graphicx}
\usepackage{subfig}
\usepackage[belowskip=0pt,aboveskip=0pt,font=small]{caption}
\aclfinalcopy 
\def\aclpaperid{2718} 

\usepackage{todonotes}

\newcommand{\xb}{\mathbf{x}}
\newcommand{\yb}{\mathbf{y}}
\newcommand{\rb}{\mathbf{r}}
\newcommand{\gb}{\mathbf{g}}
\newcommand{\ob}{\mathbf{o}}
\newcommand{\Eb}{\mathbf{E}}

\newcommand{\RR}{\mathbf{R}}
\newcommand{\GG}{\mathbf{G}}

\title{Neural Entity Recognition with Gazetteer based Fusion}

\author{Qing Sun \\
  Amazon AWS AI \\
  \texttt{qinsun@amazon.com} \\\And
  Parminder Bhatia \\
  Amazon AWS AI\\
  \texttt{parmib@amazon.com} \\}

\date{}

\begin{document}
\maketitle
\begin{abstract}
Incorporating external knowledge into Named Entity Recognition (NER) systems has been widely studied in the generic domain. In this paper, we focus on clinical domain where only limited data is accessible and interpretability is important. Recent advancement in technology and the acceleration of clinical trials has resulted in the discovery of new drugs, procedures as well as medical conditions. These factors motivate towards building robust zero-shot NER systems which can quickly adapt to new medical terminology.   We propose an auxiliary gazetteer model and fuse it with an NER system, which results in better robustness and interpretability across different clinical datasets. Our gazetteer based fusion model is data efficient, achieving +1.7 micro-$F_1$ gains on the i2b2 dataset using $20\%$ training data, and brings + 4.7 micro-$F_1$ gains on novel entity mentions never presented during training. Moreover, our fusion model is able to quickly adapt to new mentions in gazetteers without re-training and the gains from the proposed fusion model are transferable to related datasets.

\end{abstract}
\section{Introduction}
\label{sec:intro}
Named entity recognition (NER) \cite{lample-etal-2016-neural,ma2016end} aims to identify text mentions of specific entity types. In clinical domains, it's particularly useful for automatic information extraction, e.g., diagnosis information and adverse drug events, which could be applied for a variety of downstream tasks such as clinical event surveillance, decision support \cite{jin2018improving}, pharmacovigilance, and drug efficacy studies. 

We have witnessed a rapid progress on NER models using deep neural networks. However, applying them to clinical domain \cite{bhatia2019comprehend} is hard due to the following challenges: (a) accessibility of limited data, (b) discovery of new drugs, procedures and medical conditions and the (c) need for building interpretable and explainable models. Motivated by these, we attempt to incorporate external name or ontology knowledge, e.g., \emph{Remdesivir} is a \texttt{DRUG} and \emph{COVID-19} is a \texttt{Medical Condition}, into neural NER models for clinical applications.

Recent work on leveraging external knowledge can be categorized into two categories - Gazetteer embedding and Gazetteer models. Recent work has primarily focused on gazetteer embeddings. \citet{song2020improving} feed the concatenation of BERT output and gazetteer embedding into Bi-LSTM-CRF. \citet{peshterliev2020self} use self-attention over gazetteer types to enhance gazetteer embedding and then concatenate it with ELMO, char CNN and GloVe embeddings. By contrast, the basic idea of gazetteer model is to treat ontology knowledge as a new clinical modality. \citet{magnolini2019use} combine outputs of Bi-LSTM and gazetteer model and feed them into CRF layer. \citet{liu2019towards} apply hybrid semi-Markov conditional random field (HSCRF) to predict a set of candidate spans and rescore them with a pre-trained gazetteer model. 

\begin{figure*}[t]
    \centering
    \subfloat[Early fusion]{\includegraphics[trim=0pt 0pt 0pt 0pt, clip=true, width=0.45\textwidth]{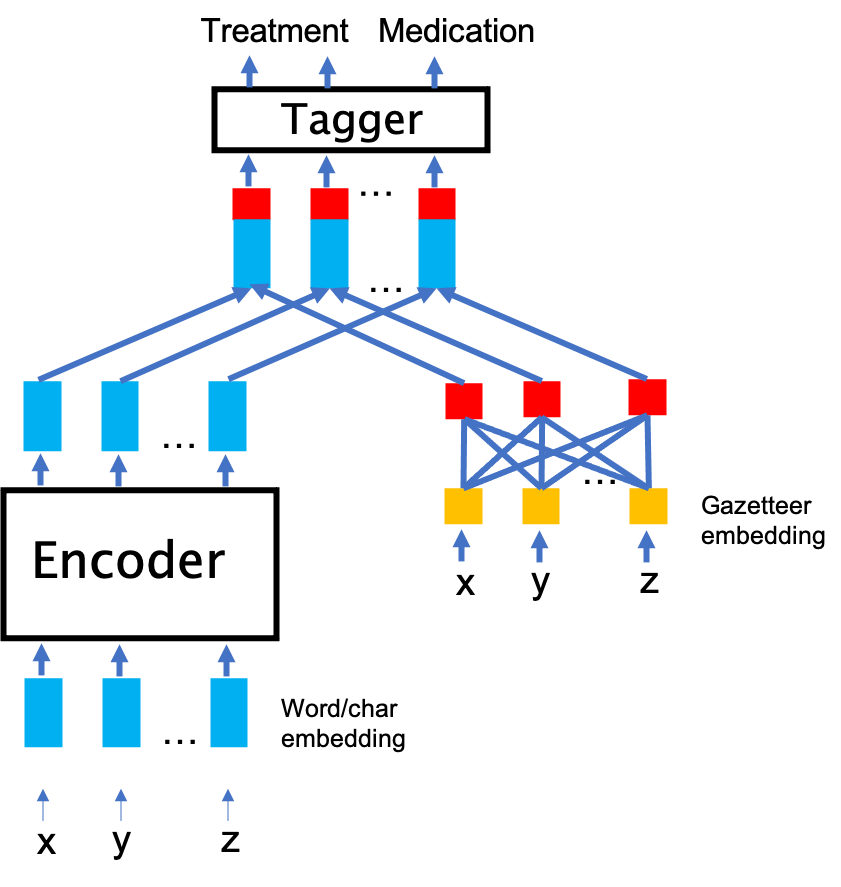} \label{fig:early}} \quad \quad
    \subfloat[Late fusion]{\includegraphics[trim=0pt 0pt 0pt 0pt, clip=true, width=0.45\textwidth]{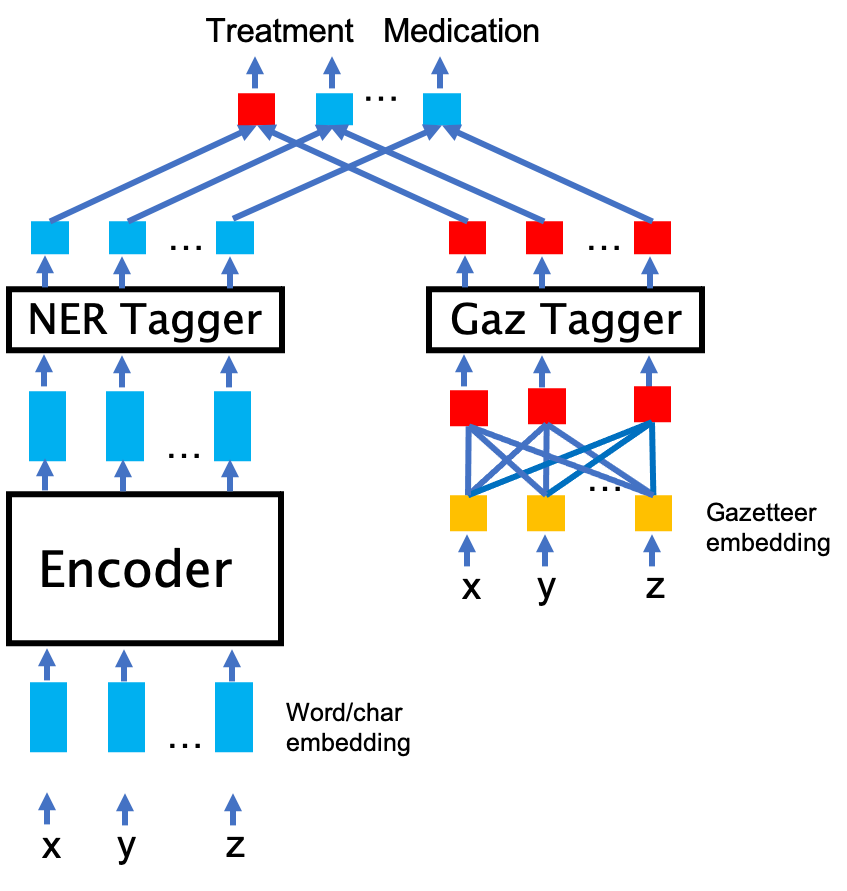} \label{fig:late}}
    \caption{Model Architecture. (a) Early fusion. The outputs of NER and gazetteer are concatenated and fed into a shared tagger. (b) Late fusion. NER and gazetteer apply separate taggers and two modalities are fused by taking element-wise max pooling.}
    \label{fig:teaser}
     \vspace{-15pt}
\end{figure*}

In this paper, we combine the advantages of both worlds. Unlike the work of \citet{peshterliev2020self}, we build self-attention over entity mentions and their context rather than over different gazetteer types. For example,  \emph{Take Tylenol 3000 \texttt{(NUM)} mg \texttt{(METRIC)} per day}, in which \emph{Tylenol} is more likely to be a \texttt{DRUG} given \texttt{NUM}, \texttt{METRIC} in context. Moreover, we study two fusion methods to integrate information from two modalities.
\vspace{-5pt}
\begin{itemize}
\setlength\itemsep{-5pt}
    \item \emph{Early fusion}. Similar to \citet{magnolini2019use}, NER model and gazetteer model apply a shared tagger, as shown in Fig.~\ref{fig:early}
    \item \emph{Late fusion}. For better interpretability and flexibility, we allow NER and gazetteer models to apply separate taggers and fuse them before taking softmax, as shown in Fig.~\ref{fig:late}
\end{itemize}

Unlike the work of \citet{liu2019towards}, NER and gazetteer models are jointly learned end-to-end.

Our contributions are as follows. (1) We propose to augment NER models with an auxiliary gazetteer model via \emph{late fusion}, which provides better interpretability and flexibility. Interestingly, the NER model can preserve the gains even if the gazetteer model is unplugged at inference time. (2) Our thorough analysis shows that the fusion model is data efficient, explainable and is able to quickly adapt to novel entity mentions in gazetteers. (3) Experiments show that the fusion model consistently brings gains cross different clinical NER datasets.

\section{Approach}

\subsection{NER model}
\label{sec:ner}
NER is a sequence tagging problem by maximizing a conditional probability of tags $\yb$ given an input sequence $\xb$. 
We first encode $\xb$ into hidden vectors and apply a tagger to produce output $\yb$. 
\setlength{\belowdisplayskip}{0pt} \setlength{\belowdisplayshortskip}{0pt}
\setlength{\abovedisplayskip}{5pt} \setlength{\abovedisplayshortskip}{0pt}
\begin{align}
 \rb & = \text{Encoder}_{R}\big(\xb \big)  \\
 \ob^{r}_t &= \text{Tagger}_{R}\big(\rb_t\big) \label{eqn:tagger} \\
 \yb_t &= softmax(\ob^{r}_t) 
\end{align}

\subsection{Gazetteer model}
We embed gazetteers into $\Eb \in \mathbb{R}^{M \times K \times d}$, where $M$ is the number of gazetteers (e.g, drugs, medical condition), $K$ is the number of gazetteer labels (e.g, B-Drug, E-Drug), and $d$ is the embedding size. We define $\Eb^g_t=[\Eb_{0, z^0_t}; \Eb_{1, z^1_t}; \cdots, \Eb_{M, z^M_t}]$, where $z^j_t$ is the gazetteer label of token $\xb_t$ in gazetteer $j$. In order to model the association of name knowledge between entity mentions and their contexts, we compute context-aware gazetteer embedding using scaled dot-product self-attention
\begin{align}
   \gb_t & = softmax\big(\frac{\Eb^g_t(\Eb^g_{t'})^T}{\sqrt{d}}\big) \Eb^g_t, \, \forall |t- t'| \leq w
\end{align}
where $w$ is the size of attention window. 

Similar to the NER model, we apply a tagger to produce output $\yb$
\begin{align}
 \ob^{g}_t &= \text{Tagger}_{G}\big(\gb_t\big) \label{eqn:tagger} \\
 \yb_t &= softmax(\ob^{g}_t) 
\end{align}

\subsection{Fusion: NER + gazetteer}
\label{sec:fusion}
To better use information from both modalities, we investigate two different fusion methods to combine information from NER and gazetteer.
\begin{itemize}
    \item \emph{Early fusion.} In Fig. \ref{fig:early}, we concatenate $\rb_t$ with $\gb_t$, and feed it into a shared tagger
\begin{align}
    \yb_t = softmax\big(\text{Tagger}_{RG}\big([\rb_t ; \gb_t]\big)\Big)
    \label{eqn:early}
\end{align}
\item \emph{Late fusion.} In Fig. \ref{fig:late}, we directly fuse $\ob^{r}_t$ and $\ob^{g}_t$ by performing element-wise max pooling
\begin{align}
    \yb_t = softmax\big(\max(\ob_t^r, \ob_t^g)\big) \label{eqn:late}
\end{align}
\end{itemize}

\section{Experiments}
\label{sec:length}

\subsection{Experimental setup}

\begin{table}
  \centering 
  \small \addtolength{\tabcolsep}{-0pt}
  \caption{Results on i2b2 (Med, TTP) and DCN (Med, DS). We report micro-$F_1$ score, each is averaged over 3 random seeds.}
  \begin{tabular}{lrrrl} 
    \toprule 
    & \multicolumn{2}{c}{i2b2} & \multicolumn{2}{c}{DCN}\\
    \cline{2-5}
     & Med & TTP & Med & DS \\
    \midrule 
    NER w/o fusion & 92.26 & 87.22 & 84.51 & 83.99\\
    \midrule
    Early fusion  & 92.14 & 87.42 & 84.82 & 84.51 \\
    Early fusion + attention & 92.44 & 87.43  & 84.99 & 84.47 \\
    \midrule
    Late fusion & 92.37 & 87.32 & 84.84 & 84.58 \\
    Late fusion + attention & 92.35 & 87.41 & 84.82 & 84.37 \\
    \bottomrule 
  \end{tabular}
   \vspace{-15pt}
  \label{tab:results}
\end{table}

\textbf{LM pre-training.} We continue to pre-train $\text{RoBERTa}_{\text{base}}$ (L=12, H=768, A=12) \cite{liu2019roberta} on MIMIC-III dataset \cite{johnson2016mimic}, which comprises deidentified clinical data from $\sim$ 60k intensive care unit admissions.

\paragraph{Fine-tuning on clinical NER datasets.} We fine-tune $\text{RoBERTa}_{\text{mimic}}$ and learn a gazetteer model (w/ NER tagger) from scratch on clinical datasets.

\begin{itemize}
\setlength\itemsep{-5pt}
    \item i2b2 -  We use public datasets from the 2009 and 2010 i2b2 challenges for medication (Med) \cite{uzuner2010extracting}, and ``test, treatment, problem'' (TTP) entity extraction. We follow the original data split from \citet{chalapathy2016bidirectional} of 170 notes for training and 256 for testing. 
    \item De-identified clinical notes (DCN) - Second dataset \cite{bhatia2018joint} consists of 1,500 de-identified, annotated clinical notes with medications (Med) and medical conditions (DS). We follow i2b2 challenge guidelines for data annotation.
\end{itemize}
\vspace{-5pt}
We extract medical condition and drug dictionaries from UMLS\cite{bodenreider2004unified} (ontology knowledge graph) based on graph as well semantic meanings. We followed different steps to prune the dictionaries based on different medical ontologies such as RxNorm for medication ($\sim$100k concepts), ICD-10 CM and SNOMED for medical conditions ($\sim$500k concepts). We employ Inside, Outside, Begin, End and Singleton (IOBES) format for both tags and gazetteers\footnote{We do string matching for gazetteers by following \cite{chiu2016named}. For example, if A, B and AB are all in gazetteers, we’ll label AB as AB. The basic idea is to start from bigger spans, so we first check for ABC, if not found then AB, if not found then A and B.}.

We minimize the cross-entropy loss during training and report micro-$F_1$ score at test time. We use $\text{RoBERTa}_{\text{mimic}}$ as NER encoder and parameterize Taggers via Multi-layer Perception (MLPs). We use BertAdam optimizer, learning rate $5e^{-5}$, and dropout $0.1$. We tune hyper-parameters $d \in [2, 12]$ (best:$8$) and $w \in [2, 10]$ (best:$5$) on validation set.

\subsection{Results.} We report overall results in Table \ref{tab:results}. We observe that incorporating name knowledge consistently boost performance on all datasets by $0.18\sim0.59$ micro-$F_1$ gains. Overall, two fusion methods achieve comparable results.

\subsection{Analysis}
We investigate the effectiveness of late fusion on handling three challenges: novel entity mentions, little data access and interpretability.

\begin{table}[t]
  \centering 
  \small \addtolength{\tabcolsep}{-0pt}
  \caption{Performance on unseen entity mentions. Models are trained using $20\%$ training data. We report performance of Medication in i2b2 Med and Treatment in i2b2 TTP.}
  
  \begin{tabular}{lrl} 
    \toprule 
     & Medication & Treatment \\
    \midrule 
    NER w/o fusion & 76.96 & 72.40\\
    Late fusion w/ attention & 81.63 \textbf{(+4.7)} & 74.30 \textbf{(+1.9)}  \\
    \bottomrule 
  \end{tabular}
  \label{tab:zero_shot}
\end{table}

\begin{table}[t]
  \centering 
  \small \addtolength{\tabcolsep}{0pt}
  \caption{Ablation study on individual modules.}
  \begin{tabular}{cccc} 
    \toprule 
      $R_0$ & $RG$ & $R$ & $R_0G$ \\
      \midrule
    76.23 & 96.33 & 90.71 & 85.75\\
    \bottomrule 
  \end{tabular}
  \label{tab:interpretability}
  \vspace{-15pt}
\end{table}

\subsubsection{Novel entity mentions}
New drugs and medical condition come out very frequently. For example,  ``remdesivir''and  ``Baricitinib'' for COVID-19. To investigate the effect of late fusion on unseen entity mentions, we focus on answering questions: whether it can generalize well on unseen entity mentions, and whether it is able to correct prediction once novel entity names are added into gazetteer without re-training?

\paragraph{Zero-shot.} We report results on unseen entity mentions not presented in train and validation sets. In Table \ref{tab:zero_shot}, we see that late fusion brings significant improvement: $+4.7$ F1 for medication (i2b2 Med) and $+1.9$ F1 for Treatment (i2b2 TTP).

\paragraph{``One''-shot in gazetteer.} We evaluate the ability of late fusion to quickly adapt to non-stationary gazetteers, e.g., specialists might add new entity mentions into gazetteers or give feedback when models make incorrect prediction. 

For this analysis, we split entity mentions in training set into two parts: $70\%$ labelled and $30\%$ in gazetteer, and compare models: 
\vspace{-5pt}
\begin{itemize}
\setlength\itemsep{0pt}
\item $\RR_0$: NER model only
\item $\RR\GG$: $\RR$ + $\GG$ via late fusion
\item $\RR$: Unplug $\GG$ from $\RR\GG$ after training
\item $\RR_0\GG$: Fix $\RR_0$ and learn $\GG$ via late fusion
\end{itemize}
\vspace{-5pt}
where $\RR$ is NER model and $\GG$ is gazetteer model. 

In Table \ref{tab:interpretability}, we observe that $\GG$ plays two roles: (1) $\RR > \RR_0$. $\GG$ can regularize $\RR$ to gain better generalization ability, and (2) $\RR_0\GG > \RR_0$ and $\RR\GG > \RR$. Besides serving as a regularizer, $\GG$ provides extra information at test time. 

\begin{figure}[t]
    \centering
    \includegraphics[trim=8pt 10pt 0pt 53pt, clip=true, width=0.35\textwidth]{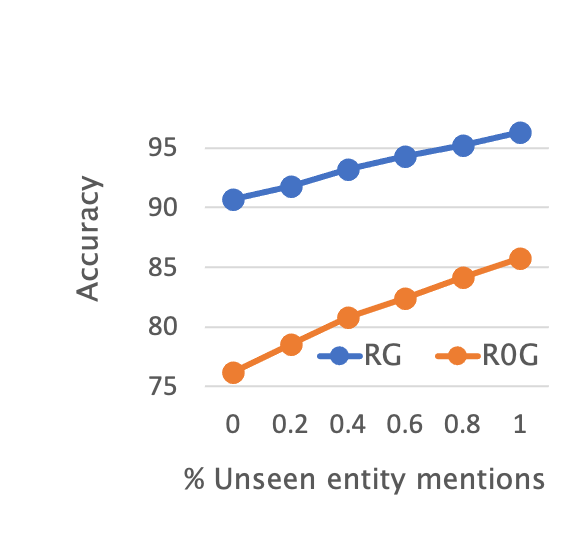}
    \caption{Quick adaptation to non-stationary gazetteers. As we increase the number of unseen entity mentions included in gazetteers, the performance goes up without re-training.}
    \label{fig:user}
    \vspace{-8pt}
\end{figure}

Moreover, we evaluate late fusion by varying the number of unseen entity mentions included in gazetteers. In Fig. \ref{fig:user}, without re-training models, late fusion can adapt to new mentions and obtain linear gains, which enables effective user feedback. 

Overall, the ability to detect and adapt to novel entity mentions, without re-training models, is useful with accelerated growth in drug development as well as in practical settings where entity extraction is one of the components to build knowledge graph and search engines \cite{wise2020covid,bhatia2020aws}. For example, linking new drugs discovered in clinical trails of COVID-19 to standardized codes in ICD-10 \footnote{\tiny \url{https://www.cdc.gov/nchs/icd/icd10cm.htm}} or SNOMED \footnote{\tiny \url{https://www.nlm.nih.gov/healthit/snomedct/index.html}}.

\begin{table}[t]
  \centering 
  \small \addtolength{\tabcolsep}{-0pt}
  \caption{Cross-evaluation on i2b2 Med and DCN Med. Column: dataset models are trained on. Row: dataset models are evaluated on.}
  \begin{tabular}{ccc} 
    \toprule 
      & i2b2 & DCN \\
    \cline{2-3}
    \multirow{2}{*}{i2b2} & 94.54 $\rightarrow$ 94.77 & 68.78 $\rightarrow$ 69.68 \\
    & (+0.23) & (+0.9) \\
    \multirow{2}{*}{DCN} & 59.98 $\rightarrow$ 60.08 & 90.02 $\rightarrow$ 90.71 \\
    & (+0.1) & (+0.69) \\
    \bottomrule 
  \end{tabular}
  \label{tab:cross}
   \vspace{-15pt}
\end{table}

\begin{figure}[t]
    \centering
    \includegraphics[trim=8pt 10pt 10pt 0pt, clip=true, width=0.35\textwidth]{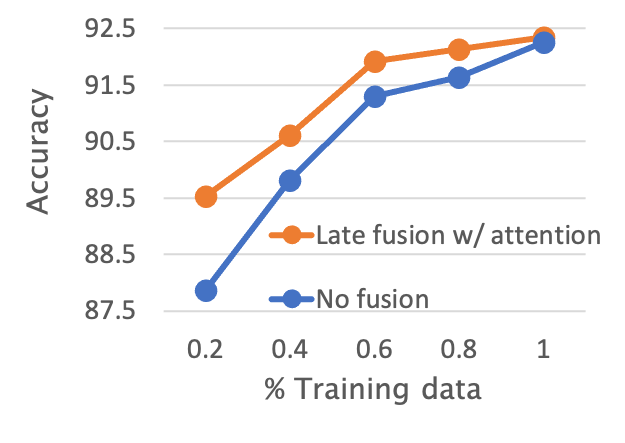}
    \caption{Accuracy vs. Training data size on i2b2 Med. We randomly sample $20\%, 40\%, \cdots, 100\%$ of training data and report micro-$F_1$ score averaged over 3 random seeds.}
    \label{fig:efficiency}
    \vspace{-8pt}
\end{figure}

\begin{table}[t]
  \centering 
  \small \addtolength{\tabcolsep}{0pt}
  \caption{Qualitative examples.}
  \begin{tabular}{clll} 
    \toprule 
  \multicolumn{2}{l}{(1) Treated for $\underbrace{\text{COPD flare}}_{\text{B-R, I-R}}$ \text{with supplemental DuoNebs}} \\
  \midrule
  NER w/o fusion & \text{B-R, O}  \\
  Late fusion w/o attention & \text{B-R, O} \\
  Late fusion w/ attention & \text{B-R, I-R} \\
 \midrule
\multicolumn{2}{l}{(2) Postop day 0, increase $\underbrace{\text{sodium}}_{\text{S-M}}$, \text{free water added}} \\
  \midrule
  \multicolumn{2}{c}{$\RR$: \text{ O}, \quad   $\GG$:\text{ S-M}, \quad $\RR\GG$:\text{ S-M}} \\
\bottomrule 
  \end{tabular}
  \label{tab:qualitative}
  \vspace{-10pt}
\end{table}


\subsubsection{Limited data access}
Typically, data accessible to use in the clinical domain is quite limited. In this section, we focus on evaluating fusion model in low-resource settings as well as investigate whether the gain is transferable across related datasets. Here we present results with late fusion methodology.  

\paragraph{Low-resource setting} We evaluate late fusion by reducing training data size from $100\%$ to $20\%$.  Fig. \ref{fig:efficiency} shows late fusion gains more when less training data is present. With $20\%$ training data, late fusion is able to boost performance over the baseline model by 1.7 micro-$F_1$ on i2b2 Med dataset.

\paragraph{Transfer learning.} To verify the generalization ability of late fusion, we train models on one dataset and report evaluation on another data source. We re-train models on i2b2 Med and DCN Med using common entity types: Dosage, Medication, Frequency, and Mode. Table.\ref{tab:cross} shows that the gains from gazetteer enhanced fusion models are preserved in i2b2 $\rightarrow$ DCN and DCN $\rightarrow$ i2b2.

\subsubsection{Interpretability}
Explainable and controllable models are very important for clinical applications. Unfortunately, it is extremely challenging for deep neural networks. We illustrate two qualitative examples in Table.\ref{tab:qualitative}. Late fusion models are trained on i2b2 Med using $20\%$ training data.
\vspace{-8pt}
\begin{itemize}
    \setlength\itemsep{-5pt}
    \item[(1)] Late fusion correctly predicts \emph{flare} as I-R (Reason) since \emph{COPD flare} is a \texttt{Medical Condition}.
    \item[(2)] By looking into individual predictions from $\RR$ and $\GG$, we notice that correct prediction is caused by name knowledge in gazetteers.
\end{itemize}
\vspace{-8pt}
Overall, late fusion provides us a tool for diagnosis system: to answer questions whether NER or gazetteer model failed and explain why mentions belong to a particular entity type.
\section{Conclusion}
We studied fusion methods to improve NER system by leveraging name knowledge from gazetteers. We did a thorough analysis on the effectiveness of fusion methods on handling limited data and non-stationary gazetteers. In addition, we demonstrated that fusion models are explainable and can be used to improve NER systems. Future research should extend our approach to structured knowledge to further improve NER system and gain better interpretability.

\section*{Acknowledgements}
We would like to thank Mohammed Khalilia and Kristjan Arumae for helpful discussions and feedback on the draft of the paper, and the anonymous reviewers for their comments.

\bibliographystyle{acl_natbib}
\bibliography{acl2021}


\end{document}